\documentclass[journal]{IEEEtran}
\usepackage{blindtext}
\usepackage{graphicx}
\usepackage{subfigure}
\usepackage{multirow}

\ifCLASSINFOpdf
\else
\fi
\begin{document}
%
\title{A Real-time Hand Gesture Recognition and Human-Computer Interaction System}
%
%
%

\author{
  \IEEEauthorblockN{Pei Xu\IEEEauthorrefmark{1}}\\
  \IEEEauthorblockA{
    Department of Electrical and Computer Engineering,\\
    University of Minnesota, Twin Cities\\
    Email: \IEEEauthorrefmark{1}xuxx0884@umn.edu
  }
}

%
%

\markboth{}%
{}
%



\maketitle

\begin{abstract}
In this project, we design a real-time human-computer interaction system based on hand gesture. The whole system consists of three components: hand detection, gesture recognition and human-computer interaction (HCI) based on recognition; and realizes the robust control of mouse and keyboard events with a higher accuracy of gesture recognition. Specifically, we use the convolutional neural network (CNN) to recognize gestures and makes it attainable to identify relatively complex gestures using only one cheap monocular camera. We introduce the Kalman filter to estimate the hand position based on which the mouse cursor control is realized in a stable and smooth way. During the HCI stage, we develop a simple strategy to avoid the false recognition caused by noises - mostly transient, false gestures, and thus to improve the reliability of interaction. The developed system is highly extendable and can be used in human-robotic or other human-machine interaction scenarios with more complex command formats rather than just mouse and keyboard events.
\end{abstract}


%
\IEEEpeerreviewmaketitle

\section{Introduction}
The hand gesture, during daily life, is a natural communication method mostly used only among people who have some difficulty in speaking or hearing. However, a human-computer interaction system based on gestures has various application scenarios. For example, play games who intend to provide mouse- and keyboard-free experience; control robots in some environment, e.g. the environment underwater, where it is inconvenient to use speech or typical input devices; or simply provide translation for people who use different gesture languages.

The premise of implementing such a system is gesture recognition. Gesture recognition has become a hot topic for decades.  Nowadays two methods are used primarily to perform gesture recognition. One is based on professional, wearable electromagnetic devices, like special gloves. The other one utilizes computer vision. The former one is mainly used in the film industry. It performs well but is costly and unusable in some environment. The latter one involves image processing. However, the performance of gesture recognition directly based on the features extracted by image processing is relatively limited. Although the performance has improved as the appearance of advanced sensors, like Microsoft Kinect sensors, the relatively higher price of such devices is still an obstacle to the large-scale application of gesture-based HCI systems. Besides, such advanced sensors perform even more unreliably than optical cameras in some certain environment. For instance, the attenuation of infrared ray in water could largely limit the use of those like Microsoft Kinect sensors in water with a good light condition.

The nature of gesture recognition is a classification problem. There are lots of approaches to handle 2D gesture recognition, including the orientation histogram~\cite{IEEEhowto:gesture_hog}, the hidden Markov model~\cite{IEEEhowto:gesture_markov}, particle filtering~\cite{IEEEhowto:gesture_particle}, support vector machine (SVM)~\cite{IEEEhowto:gesture_svm}, etc.~\cite{IEEEhowto:gesture_collection1}\cite{IEEEhowto:gesture_collection2}. Most of those approaches need preprocessing the input gesture image to extract features. The performance of those approaches depends a lot on the feature learning process. Recently, with the development of hardware, much research about object recognition using CNNs becomes practical and achieves success~\cite{IEEEhowto:app_cnn1}\cite{IEEEhowto:app_cnn2}\cite{IEEEhowto:app_cnn3}. CNNs usually learn features directly from input data and often provide a better classification result in the case where features are hard to be extracted directly, such as image classification. In this project, we design a real-time, gesture-based HCI system who employs a CNN to learn features and to recognize gestures only using one cheap monocular camera. We simplify the process of recognition in order to make the system usable on the platform with limited hardware resources when keeping the high accuracy and reliability. We train a model with a gesture set composed of 16 kinds of static gestures. The accuracy rate of the recognition can reach over 99.8\%. Moreover, due to that we use a CNN with a simple structure, the classification process can be done quite fast and thus makes the whole system runnable in real time.

Another challenge this project deals with is the implementation of mouse cursor control. Although mouse cursor control is not a must-have function in some application scenarios like human-robotic interaction, it is necessary in some other scenarios, like gaming or operating a device with a complex graphic user interface. It is expected that the mouse cursor can be controlled by hand gestures in a smooth and stable way with acceptable sensitivity. However, the hand itself is not a reliable mean to control the mouse cursor. The hand itself, due to the variability of gestures, is unable to provide a stable, easily found point, which should always be able to be tracked. Besides, it is quite hard for a person to strictly keep his hand stable when controlling the mouse cursor, especially when changing gestures to realize different control effects (e.g. changing gestures to control the mouse cursor from `move' to `left click'). In order to solve these problems, some gesture control schemes introduce extra wearable or portable devices, like colored tips or light balls, as a mark, through tracking which the mouse cursor control can be implemented. In this project, we impose some limitation on the set of gestures, who can implement mouse cursor control, and then make it simple to position a continuously trackable point on the hand without any extra mark. Furthermore, we introduce the Kalman filter to further improve the stability and smoothness of the motion of mouse cursor controlled by the hand. The main purpose of the Kalman filter here is to limit the motion of mouse cursor, prevent the cursor from jumping on the screen with the movement of the tracked hand, and thus make its motion smooth.

As what we mention above, the hand is an unstable object. The change from one gesture to another would cause a series of transient, intermediate gestures, to whom the HCI system should not make any response. In order to improve the reliability of the system, we propose a simple strategy to control the behavior of the system and prevent the system from responding to the gesture signal immediately when a new gesture signal is received from the recognition process. A more reliable scheme is proposed in~\cite{IEEEhowto:gesture_confirm}, who introduces a feedback mechanism to make confirmation with the human operator when a signal, which leads to a command with high cost or low likelihood, is received. That scheme is suitable in the scenario where there is no high demand on the response speed of the interaction system and where the likelihood and cost of a command are predictable.

Basically, the gesture-based HCI system developed in this project follows three principals: real-time, reliable and low-cost. While the system is originally developed to post mouse and keyboard events to the x86 computer based on hand gesture recognition, it is quite easy to extend the system and make it able to carry out more complex commands. As a proof and additional work, we make a little modification to the system and change the system into an interface to control autonomous robots remotely.

\section{Background and Related Work}
The gesture recognition and HCI system developed in this project involves a set of problems, mainly including hand detection and background removal, gesture recognition, mouse cursor control by hand gestures and behavior control of the system.

Hand detection and background removal are indispensable to gesture recognition. We need to segment the hand region from the background so that the gesture recognition algorithm can work properly. Some gesture recognition methods~\cite{IEEEhowto:gesture_markov}\cite{IEEEhowto:gesture_svm}\cite{IEEEhowto:gesture_recognition1} simply bypass this problem by assuming that the background can be filtered out easily. However, hand detection and background removal using only one monocular camera could become a complex problem in the practical scenario. A disadvantage of using one monocular camera is the lack of depth information such that it becomes a problem to separate the hand region from the background. Many methods are proposed to perform hand detection. Most of them are based on shape~\cite{IEEEhowto:hand_shape}, color~\cite{IEEEhowto:hand_color}, Harr features~\cite{IEEEhowto:hand_haar} or context information~\cite{IEEEhowto:hand_context}. Most of these methods work well if we impose some limitation on the environment from which the detector finds out the hand. However, all of these methods have their limitations in practical scenarios where the background environment may be cluttered, changeable and unpredictable. For examples, color-based methods may become useless when people wear gloves or when the color or mixture of the background is too close to that of the hand; the methods using context information need some other parts (usually face) of the human are visible, which also may be unrealizable in some scenarios; and shape- and Harr feature-based methods may fail when facing unforeseen hand gestures. In our developed system, we leave an interface for users to calibrate the hand detector by applying the color filter and background subtraction. This scheme is not a grandmaster key to all possible application scenarios but can deal with the case with an unchangeable or distinct background.

\begin{figure}[!t]
	\centering
	\subfigure[]{
		\includegraphics[width=0.7in]{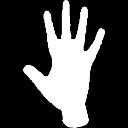}
	}
	\subfigure[]{
		\includegraphics[width=0.7in]{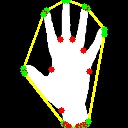}
	}
	\subfigure[]{
		\includegraphics[width=0.7in]{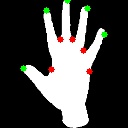}
	}
	\caption{An example of detecting fingers using convexity defects. (a) is an binary image of a palm. (b) shows the detected convexity defects. The green points are the start or end points of the defects. The red points are the farthest points of the defects. The yellow line is the detected convex hull. (c) shows the detected fingertips after filtering invalid convexity defects.}
	\label{fig:conv_defects}
\end{figure}

The data that one monocular camera can provide are a series of sequential, static 2D images. An intuitive method for gesture recognition using static 2D images is based on convexity defects or curvature. Fig.~\ref{fig:conv_defects} shows the convexity defects detected from an binary image of a palm using the algorithm in~\cite{IEEEhowto:contour_alg} and~\cite{IEEEhowto:convex_hull_alg}. The method based on curvature is quite similar to that based on convexity defects. The nature of these methods is to detect fingers. In order to make this method effect, a model is needed to filter invalid convexity defects and to ensure that each defect represents a finger. Two common parameters, which are used to build the model, are the length and intersection angle of each defect. A more complex model can be built up if considering the eccentricity~\cite{IEEEhowto:gesture_eccentricity}~\cite{IEEEhowto:gesture_markov}, elongatedness~\cite{IEEEhowto:gesture_recognition1}, or other features that can be extracted directly from the contour information of the hand image~\cite{IEEEhowto:gesture_hog}. However, there still lacks a general model to describe hand gestures directly based on these contour features. The empirical model does not always work satisfactorily during our testing even in the case where we only count fingers. ~\cite{IEEEhowto:gesture_recognition1} uses a simple artificial neural network to identify gestures based on a set of combined contour features. \cite{IEEEhowto:gesture_svm} uses SIFT to describe and SVM to recognize gestures. \cite{IEEEhowto:gesture_motion} uses kinematic features to identify American sign language.  All of these methods cannot provide a satisfactory accuracy. Some of them even cannot reach 95\%.

\begin{figure}[!t]
	\centering
	\includegraphics[width=1.4in]{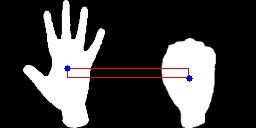}
	\caption{Relatively stable though the whole hand keeps, the position of estimated hand centers (blue points) changes significantly when the hand gesture changes from `palm' to `fist'.}
	\label{fig:center_comp}
\end{figure}

In order to conduct HCI through the mouse cursor controlled by hand gestures, a stable point must be tracked on the hand, a point based on which we can position the mouse cursor on the screen. Some papers announce the achievement of mouse cursor control only by tracking the position of the whole hand. However, hands are agile, while gestures are variable. The variability of hand gestures might lead to a significant change in the shape of the hand. Therefore, we must define a special point on the hand, a point which can be tracked stably. Why not the hand center? A direct answer is that we simply do not know where is the ground truth position of the hand center. Our estimation of the hand center is likely to vary with the change of gestures, even if the hand is kept stable during the process. Fig.~\ref{fig:center_comp} shows the shift of the estimated hand center when the gesture changes from `palm' to `fist'. A simple solution to this problem is to use wearable devices, such as colored tips~\cite{IEEEhowto:cursor_control_color1}\cite{IEEEhowto:cursor_control_color2}\cite{IEEEhowto:cursor_control_color3}. \cite{IEEEhowto:cursor_control_multi1} and \cite{IEEEhowto:cursor_control_multi2} achieve mouse cursor control through the cooperation of two hands. However, both of the two interaction methods are not natural. Besides, the screen and camera usually have different resolutions, which means that the mouse cursor cannot reach all possible pixel points on the screen if we simply linearly position the cursor on the screen based on the movement of a tracked point on the hand. Furthermore, due to the difference in the resolution, a slight movement of hands could lead to a jump of the cursor on the screen. The experience of such a mouse cursor control effect is annoying. Nevertheless, few of papers consider the smoothness of the motion of the mouse cursor controlled by hand gestures.

\begin{figure}[!t]
	\centering
	\includegraphics[width=3.5in]{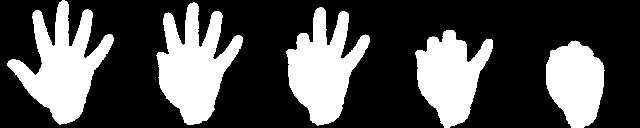}
	\caption{Some transient, false gestures are detected during the process of the gesture changing from `palm' to `fist'.}
	\label{fig:trans_gesture}
\end{figure}

The robustness of the gesture-based interaction system is not only determined by the accuracy of gesture recognition, but also the effectiveness of rejecting noise. The noise here is not necessarily caused by the cluttered background, camera blurring or some other external factors, but often is caused by the hand gesture itself and is ineluctable. Fig.~\ref{fig:trans_gesture} shows the process of the gesture changing from `palm' to `fist'. In the process, some transient gestures are detected. We called these gestures the false gestures, since we simply expect the interaction system not to make any response to these gestures. A decision scheme should be proposed to deal with the transient, false gestures. Most discussion about the similar topic is in the scenario of human-robotic interaction, where some risky actions can be taken by the robot only when the riskiness has been fully evaluated~\cite{IEEEhowto:gesture_confirm}\cite{IEEEhowto:control_safe1}\cite{IEEEhowto:control_safe2}.

\section{Methodology}
The developed gestured-based HCI system runs with fixed frames per second. Every time an image is captured by the camera, the system works as the following:
\begin{enumerate}
	\item The captured image is preprocessed and a hand detector tries to filter out the hand image from the captured image; the whole process terminates if there is nothing detected.
	\item A CNN classifier is employed to recognize gestures from the processed image, while a Kalman estimator is employed to estimate the position of the mouse cursor according to the movement of a point tracked by the hand detector.
	\item The recognition and estimation results are submitted to a control center; a simple probabilistic model is used to decide what response the system should make.
\end{enumerate}

\subsection{Hand Detection and Gesture Recognition}

\begin{figure*}[!t]
	\centering
	\includegraphics[width=7in]{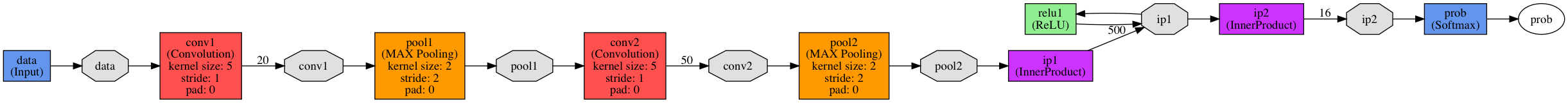}
	\caption{The structure of the CNN classifier.}
	\label{fig:net}
\end{figure*}

\begin{figure*}[!t]
	\centering
	\includegraphics[height=0.7in]{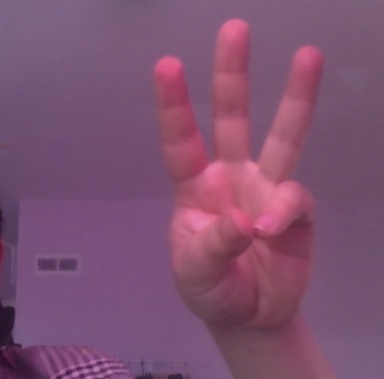}
	\includegraphics[height=0.7in]{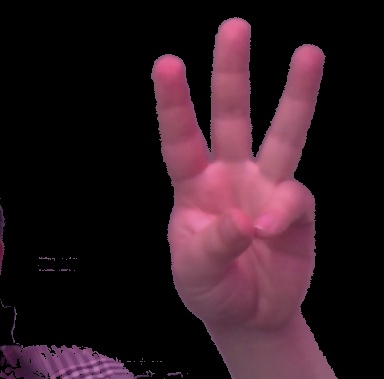}
	\includegraphics[height=0.7in]{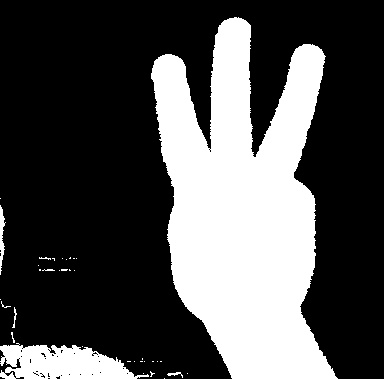}
	\includegraphics[height=0.7in]{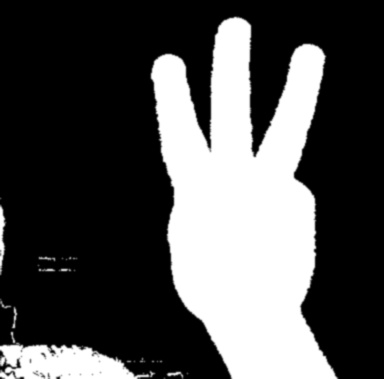}
	\includegraphics[height=0.7in]{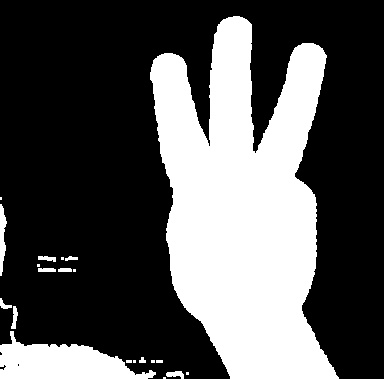}
	\includegraphics[height=0.7in]{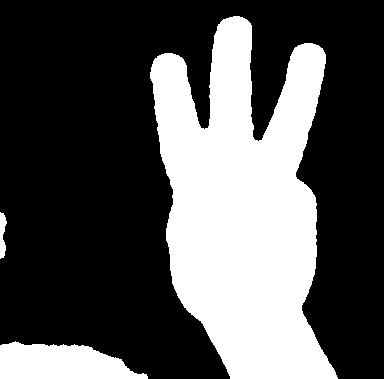}
	\includegraphics[height=0.7in]{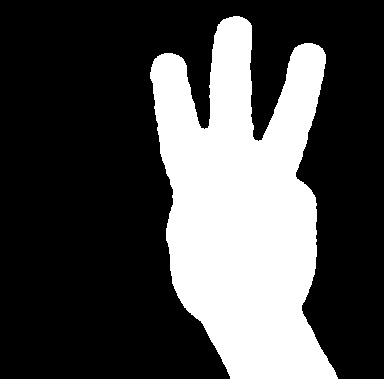}
	\includegraphics[height=0.58in]{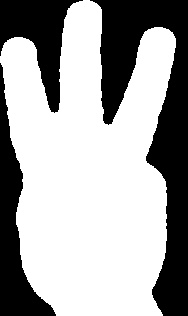}
	\includegraphics[height=0.24in]{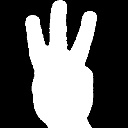}
	\caption{The process of generating the hand image, which is fed to the CNN classifier, from the image captured by a camera. The process includes background subtraction, color filtering, Gaussian blurring, thresholding, morphological transformation, contour extraction, hand region extraction, image resizing.}
	\label{fig:img_extract}
\end{figure*}

We use a CNN modified from LeNet-5~\cite{IEEEhowto:lenet} to recognize gestures. The structure of the CNN is shown in Fig.~\ref{fig:net}. The CNN feeds on a binary image such that the classifier could not be impacted by the color features of the hand. Images need to be preprocessed before fed to the CNN classifier. The preprocessing includes background subtraction (optional), hand color filtering, Gaussian blurring, thresholding, morphological transformation (opening and closing), contour extraction. A hand detector works after the preprocessing. It tracks the movement of a special point on the hand -- the point which is used to control the mouse cursor later, estimates the hand center and palm radius, and extracts the contour region of the hand. Finally, the contour region is resized into a fixed size while keeping the aspect ratio, and then is centered on a canvas, before fed to the CNN classifier. Fig.~\ref{fig:img_extract} shows the whole process of generating the image, which is fed to the CNN classifier.

\begin{figure}[!t]
	\centering
	\subfigure[]{
		\includegraphics[height=0.7in]{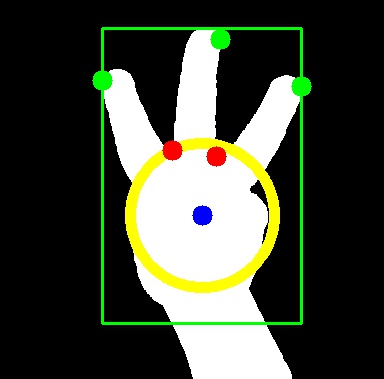}
	}
	\subfigure[]{
		\includegraphics[height=0.44in]{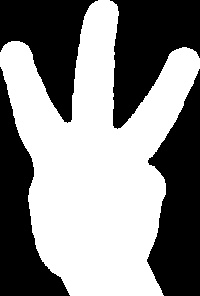}
	}
	\subfigure[]{
		\includegraphics[width=0.24in]{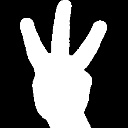}
	}
	\caption{A demonstration of how the hand region is extracted. (a) is the extracted contour region. In (a), the blue point is the estimated hand center, the green points are the detected fingertips, the red points are the farthest points of the valid convexity defects, and the yellow circle is drawn according to the estimated palm radius. Only the region enclosed by the green rectangle is kept and then (b) is obtained. Finally, (b) is resized while keeping the aspect ratio and is centered at a new canvas such that we get (c), which will be fed to the CNN classifier.}
	\label{fig:radius_est}
\end{figure}

The hand center is estimated by the distance transformation. The pixel point in the hand contour region with the maximal value after distance transformation is considered as the hand center. The palm radius is estimated by the farthest distance between the hand center and the farthest point (as the red points shown in Fig.~\ref{fig:conv_defects}) of the valid convexity defect. The word `valid' means that the convexity defect has been validated such that it is likely represents a fingertip. If there is no valid convexity defect, e.g. in the case where the gestures is a fist, the palm radius is estimated by the minimum distance between the hand center and a point outside the region of the hand contour.  The palm radius is used to estimate the location of the wrist, Then the hand region is separated from the arm based on the location of the wrist. Fig.~\ref{fig:radius_est} gives a demonstration of how the palm radius is estimated and how the arm region is separated out.

\begin{figure}[!t]
	\centering
	\subfigure[]{
		\includegraphics[width=0.7in]{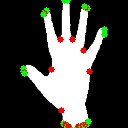}
	}
	\subfigure[]{
		\includegraphics[width=0.7in]{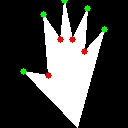}
	}
	\caption{A comparison of detecting convexity defects from the hand contour with and without polygonal approximation. The green and red points are the detected convexity defects. (a) Directly detect convexity defects without the polygonal approximation. (b) Apply polygonal approximation before detecting convexity defects.}
	\label{fig:approx_poly}
\end{figure}

In order to improve the performance of detecting convexity defects, the hand contour, before detecting convexity defects, is approximated by a polygonal curve such that each fingertip becomes a triangle-like shape, as shown in Fig.~\ref{fig:approx_poly}. The Ramer-Douglas-Peucker algorithm is employed to perform the polygonal approximation.

In the stage of gesture recognition, the CNN classifier feeds on the processed, binary images, where the hand contour is centered and adjusted to a fixed size, and then produces a probabilistic result. The system recognizes the gesture as the one with the highest probability. An advantage of employing a CNN to perform image classification is that we do not need to extract features manually. All features are extracted, or say learned, by the CNN itself. Such a characteristic often leads to a better classification result when we lack an effective mean to extract features. The CNN, which we use here to recognize gestures, contains two convolutional layers, each of which is followed by a max-pooling layer, and two fully connected layers. It uses rectified liner unit (ReLU) as the activation. Our preprocessing steps threshold, resize and center the hand image and thus introduce contrast, scale and translation invariants to some degree during the CNN learns features. The usage of max-pooling layers makes features learned by the CNN classifier be rotation-invariant to a certain extent~\cite{IEEEhowto:rotate_invariant} as well. The structure of the CNN used here is simple, at least compared to AlexNet~\cite{IEEEhowto:alexnet}, GoogLeNet~\cite{IEEEhowto:googlenet}, ResNet~\cite{IEEEhowto:resnet} or some other CNNs or deep CNNs proposed in recent years. The simple structure ensures that the recognition process can be done in real time.

\subsection{Interaction Scheme with Human}
The interaction scheme involves two problems: how to control the mouse cursor by hand gestures and how to avoid responding to the transient and false gestures as shown in Fig.~\ref{fig:trans_gesture}.

\begin{figure}[!t]
	\centering
	\subfigure[]{
		\includegraphics[width=3.5in]{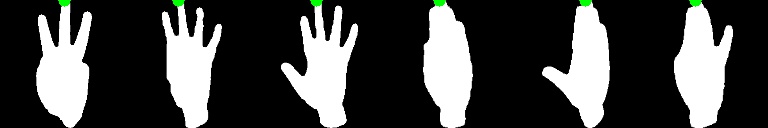}
	}
	\subfigure[]{
		\includegraphics[width=3.5in]{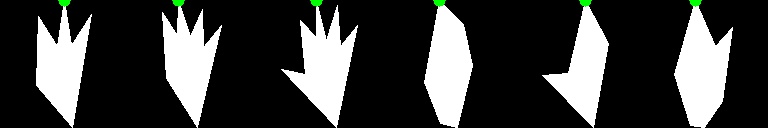}
	}
	\caption{Gestures suggested for controlling the mouse cursor. (b) is obtained by performing the polygonal approximation on (a). The green points are the suggested points for tracking. They are located from the images in (b).}
	\label{fig:gesture_mouse}
\end{figure}

The key point of the former problem is that we must, form the hand, find a trackable point through tracking which the movement of the mouse cursor can be controlled stably. Fig.~\ref{fig:gesture_mouse} shows some gestures that we suggest for controlling the mouse cursor. The characteristic of these gestures is that they are all palm-based gestures. When making transformation among these gestures, the hand can keep relatively stable. The point that we suggest for tracking is the topmost point of these palm-based gestures. Typically, the topmost point of these gestures is also the tip of the middle finger. This fingertip can be located easily especially after the polygonal approximation is applied. People usually can smoothly make change among these gestures, while keeping their middle finger immobile. Such palm-based gestures are diverse enough to support various mouse events, including `move', `single click', `double click', `right click' and `drag'.

In order to avoid responding to the transient and false gestures, we introduce a probabilistic model. Suppose that the system receives a series of gestures from the CNN classifier, $\mathcal{G}$, during a response period, i.e.
\[
	\mathcal{G} = \{g_{1}, g_{2}, \ldots, g_{n}\}
\]
where $g_{i}$ is the $i$-th gesture recognized by the system. The discriminant function to execute the command $c_{i}$, which is corresponding to a gesture $g$, is defined as
\[
	f_{i}(\mathcal{G}) = \Pr(c_{i}\vert \mathcal{G})
\]

We can introduce a Markov model or Bayes risk estimator to evaluate $f_{i}$. However, we may only have insufficient or even completely have no knowledge about the risk and appearance frequency of each gesture and command. In this case, we can simplify the discriminant function into the following form and thus make the probabilistic model work.
\[
	f_{i}(\mathcal{G}) = \frac{\sum_{j}r_{i,j}}{\vert\mathcal{G}\vert}
\]
where
\[
	r_{i,j} = \left\{\begin{array}{cc} 1, & g_{j} = c_{i} \\ 0, & g_{j} \neq c_{i} \end{array}\right.
\]
$j = 1, \ldots, n$ and $\vert\cdot\vert$ is the cardinality operation.

In the above discussion, we introduce a concept of `response period'. The developed system does not immediately make a response right after it receives a gesture recognized by the CNN classifier, but executes a command according to the discriminant function, which is computed via the sequential gestures appearing during a period. This method can effectively improve the robustness of the HCI system, but also could lead to a delay in response. The delay may interrupt some durative control effects, or say `held-on' actions. For instance, the `drag' action of the mouse cursor. The system must keep the cursor executing the `drag' action rather than do `drag' every response period, in order to prevent the `drag' action from terminating accidentally. A mean to handle this problem is that we force the system to keep executing the `held-on' action once the action is confirmed and until a new action is confirmed by the response scheme discussed above.

\section{Results}
\subsection{Gesture Recognition}

\begin{figure}[!t]
	\centering
	\includegraphics[width=3.5in]{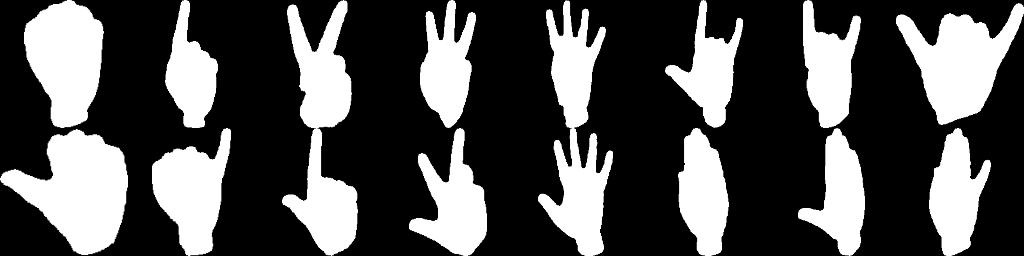}
	\caption{The gesture set collected for training. Their names are: Fist, One, V, W, Four, ILY, Rock, Loose, Thumb-Left, One-Right, Two-Left, Three-Left, Palm, Palm-Tight, Palm-Left and Palm-Right.}
	\label{fig:gesture_set}
\end{figure}

We, using the CNN, train a model including a set of 16 kinds of gestures. 19,852 sample images, in total, are collected from five people. Each gesture has more than 1,200 samples. For each gesture, 200 samples are used for validation, while the rest are used for training. Fig.~\ref{fig:gesture_set} shows the 16 kinds of gestures.

The stochastic gradient decent method is used for training. At each iteration, the estimated parameters, $\theta$, are updated via
\[
	\mathbf{\theta}^{(k+1)} = \mathbf{\theta}^{(k)} + \mathbf{v}^{(k+1)}
\]
where
\begin{equation}\label{eqn:sgd_step_update}
	\mathbf{v}^{(k+1)} = \mu \mathbf{v}^{(k)} - \alpha \nabla_{\mathbf{\theta}} L
\end{equation}
in which $\alpha$ is the base learning rate, $\mu$ is the momentum, and $L$ is the loss function.

\begin{figure}[!t]
	\centering
	\includegraphics[width=3in]{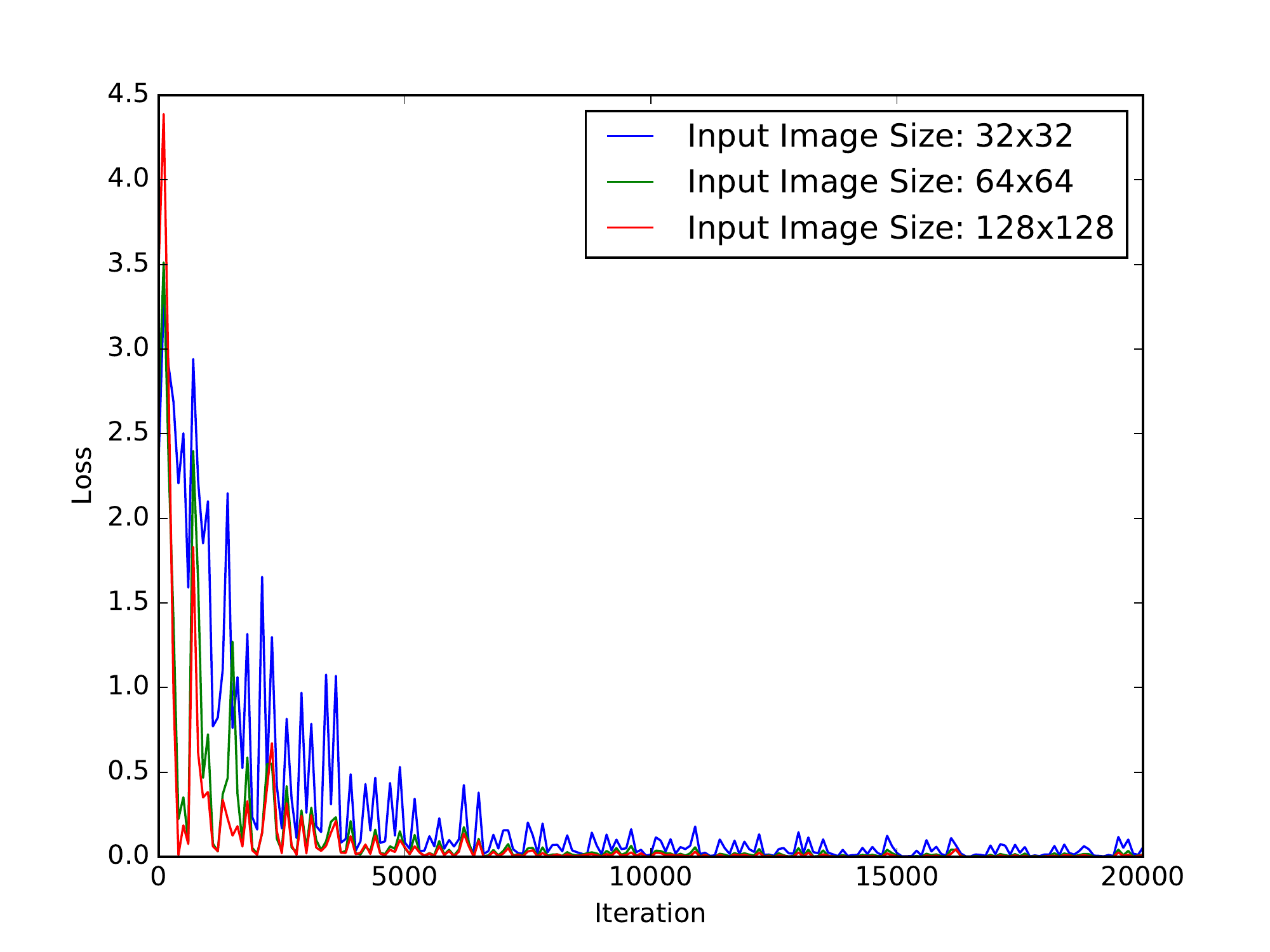}
	\caption{Loss of three models during training.}
	\label{fig:train_loss}
\end{figure}

\begin{figure}[!t]
	\centering
	\includegraphics[width=3in]{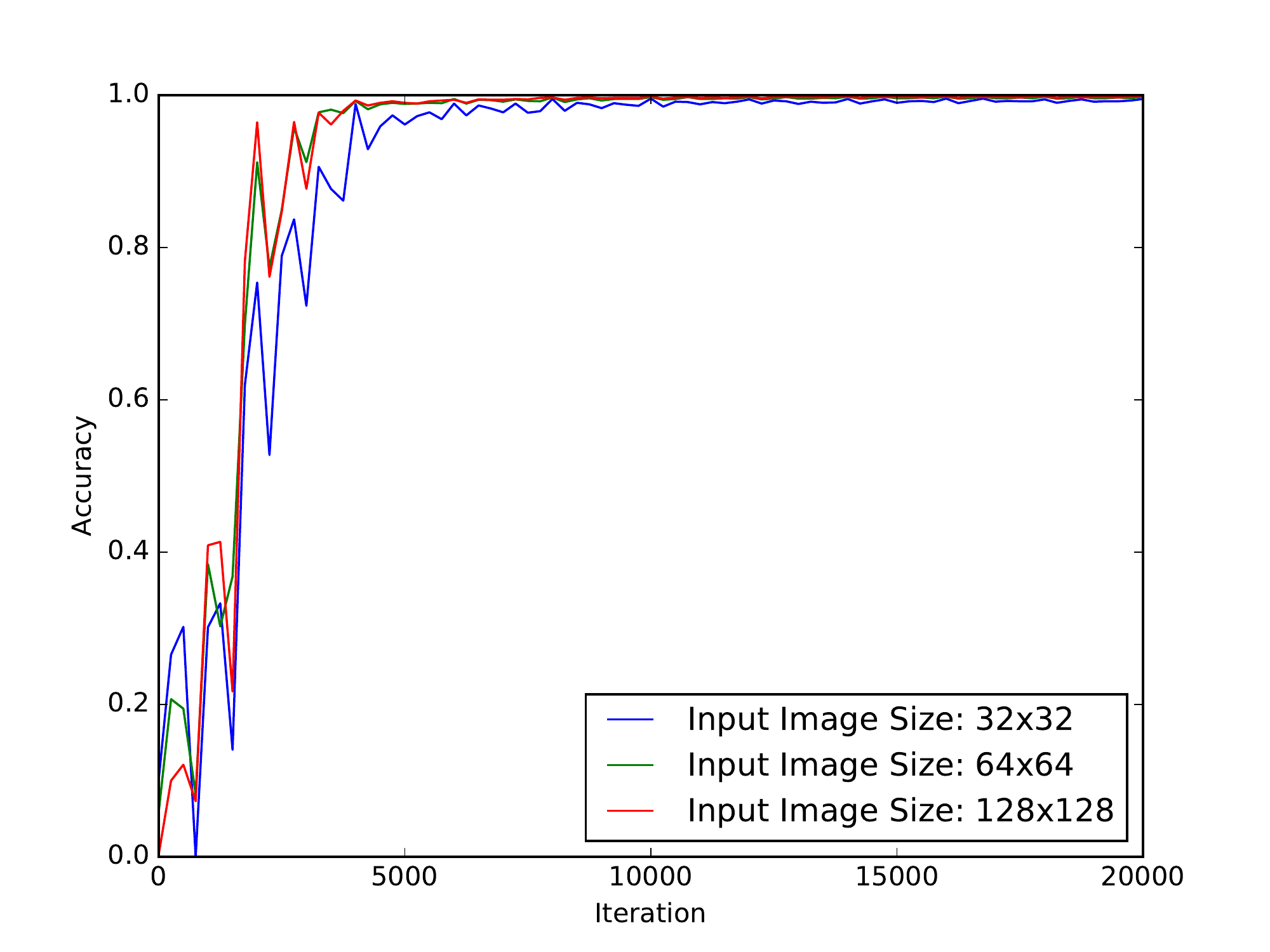}
	\caption{Accuracy of three models during testing.}
	\label{fig:test_accuracy}
\end{figure}

We train several models using sample images with the size $128\times128$, $64\times64$ and $32\times32$ respectively combined with various values of learning rate, $\alpha$, and momentum, $\mu$, and find that the model using sample images with the size $64\times64$ provides the best performance when $\alpha = 0.0001$ and $\mu = 0.9$. Fig.~\ref{fig:train_loss} shows the loss during training when $\alpha = 0.0001$ and $\mu = 0.9$. Fig.~\ref{fig:test_accuracy} shows the accuracy when applying the model to the test set during training. Basically, the model using the input image with size $128\times128$ and that using the image with size $64\time64$ can reach quite similar loss and accuracy. However, the smaller size of the input image leads to a huge reduce in computational cost. When the model using the input image with size $64\times64$ is used, the final accuracy, during our test, can reach over 99.8\%.

\subsection{Interaction Performance Evaluation}

\begin{figure}[!t]
	\centering
	\includegraphics[width=3in]{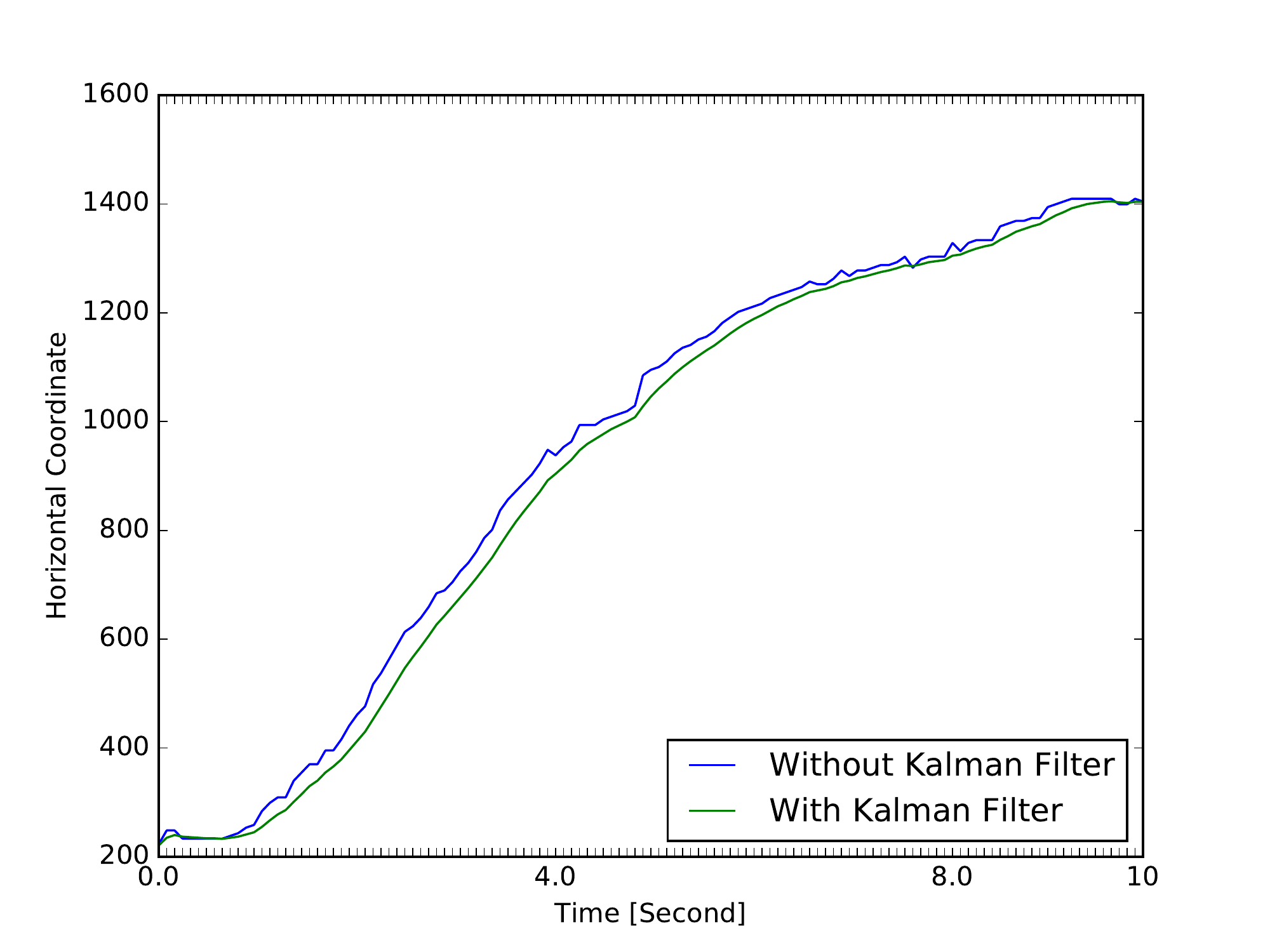}
	\caption{Change in the horizontal coordinates of the mouse cursor controlled by a hand moving horizontally from the left to right.}
	\label{fig:kalman}
\end{figure}

The Kalman filter, which is used to smooth the motion of mouse cursor, is initialized with four state variables and two measurement variables. The four state variables respectively represent the horizontal coordinate, $x$, the vertical coordinate, $y$, the horizontal offset, $\mathrm{d} x$ and the vertical offset, $\mathrm{d} y$. The two measurement variables represent the observation of the horizontal and vertical coordinate. The transformation matrix is, therefore, defined as
\[
	F = \left[\begin{array}{cccc}
	1 & 0 & 1 & 0\\
	0 & 1 & 0 & 1\\
	0 & 0 & 1 & 0\\
	0 & 0 & 0 & 1\\
	\end{array}\right]
\]

Fig.~\ref{fig:kalman} displays the change of horizontal coordinates when a mouse cursor is controlled by a hand moving from the left to right horizontally with and without the Kalman filter. We can find that the introduction of the Kalman filter makes the mouse cursor move more smoothly. Although the Kalman filter also introduces some delay, such delay is acceptable.

In order to evaluate the performance of the proposed response scheme that is based on the probabilistic model, we record the response of the system when it faces certain transformation of gestures. The results are shown in Table~\ref{tab:response_count}.  From the results, we can find that the system does not respond to transient, false gestures when gestures change and that the system barely accidentally interrupts `held-on' actions. It demonstrates that the proposed response scheme can effectively prevent the system from responding to the transient, false gestures, and, meanwhile, keep sensitive to the durative commands.

\begin{table}[!t]
	\renewcommand{\arraystretch}{1.3}
	\caption{Performance of Proposed Response Scheme}
	\label{tab:response_count}
	\centering
	\begin{tabular}{|c|c|c|c|c|c|}
		\hline
		\multicolumn{2}{|c|}{\bfseries Initial} & \multicolumn{2}{|c|}{\bfseries Final} & \multirow{2}{*}{\bfseries S$^a$} & \multirow{2}{*}{\bfseries F$^b$}\\
		\cline{1-4}
		\bfseries Gesture & \bfseries Action & \bfseries Gesture & \bfseries Action &  & \\
		\hline
		\hline
		Four & Drag (Mouse) & Four & Drag (Mouse) & 48 & 2\\
		\hline
		Palm-Tight & Move (Mouse) & Palm-Left& Click (Mouse) & 50 & 0 \\
		\hline
		Palm-Tight & Move (Mouse) & One & Up (Keyboard) & 50 & 0 \\
		\hline
		One & Up (Keyboard) & One & Up (Keyboard) & 50 & 0\\
		\hline
		One & Up (Keyboard) & Thumb-Left & Left (Keyboard) & 49 & 1\\
		\hline
	\end{tabular}\\
	\footnotesize{$^a$ S for success. $^b$ F for failure. For durative commands or `held-on' actions, such as the `drag' action, F means an accidental termination.}\\
\end{table}

\begin{figure}[!t]
	\centering
	\subfigure[]{
		\includegraphics[width=1.63in]{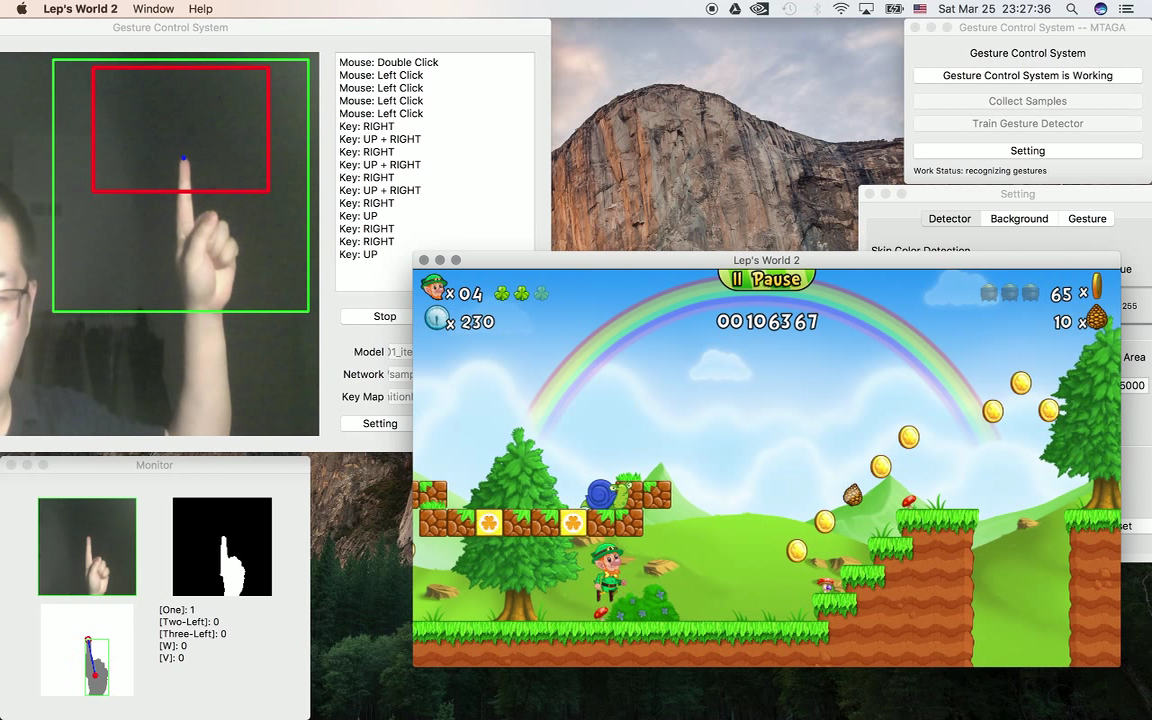}
	}
	\subfigure[]{
		\includegraphics[width=1.63in]{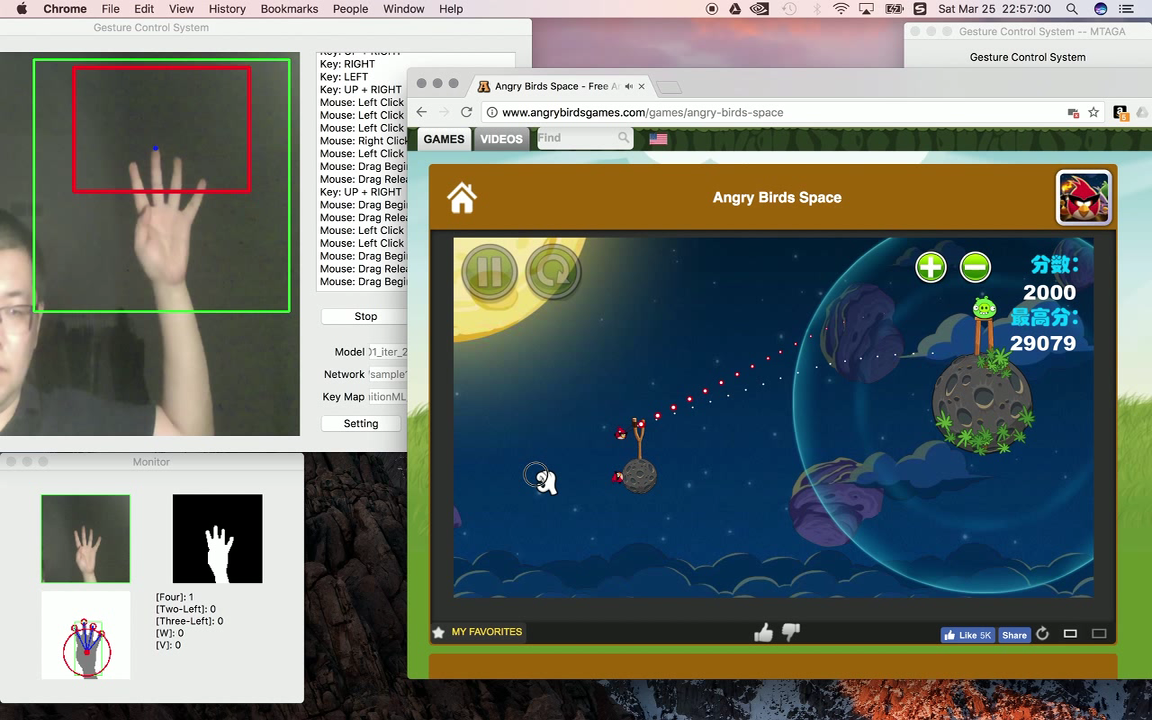}
	}
	\caption{Two demos of the developed gesture-based HCI system. In (a), the user utilizes hand gestures to trigger keyboard events and play the game \textit{Lep's World}. In (b), the user utilizes hand gestures to trigger mouse events and play the game \textit{Angry Birds}.}
	\label{fig:demo}
\end{figure}

Fig.~\ref{fig:demo} shows two demos of the developed gesture-based HCI system. In the first demo, the user utilizes the gesture `one' to command the HCI system to simulate the event caused by pressing the `up' key. In the second, the user utilizes the gesture `four' to command the system to simulate the `drag' event of the mouse.

\begin{figure}[!t]
	\centering
	\includegraphics[width=1.63in]{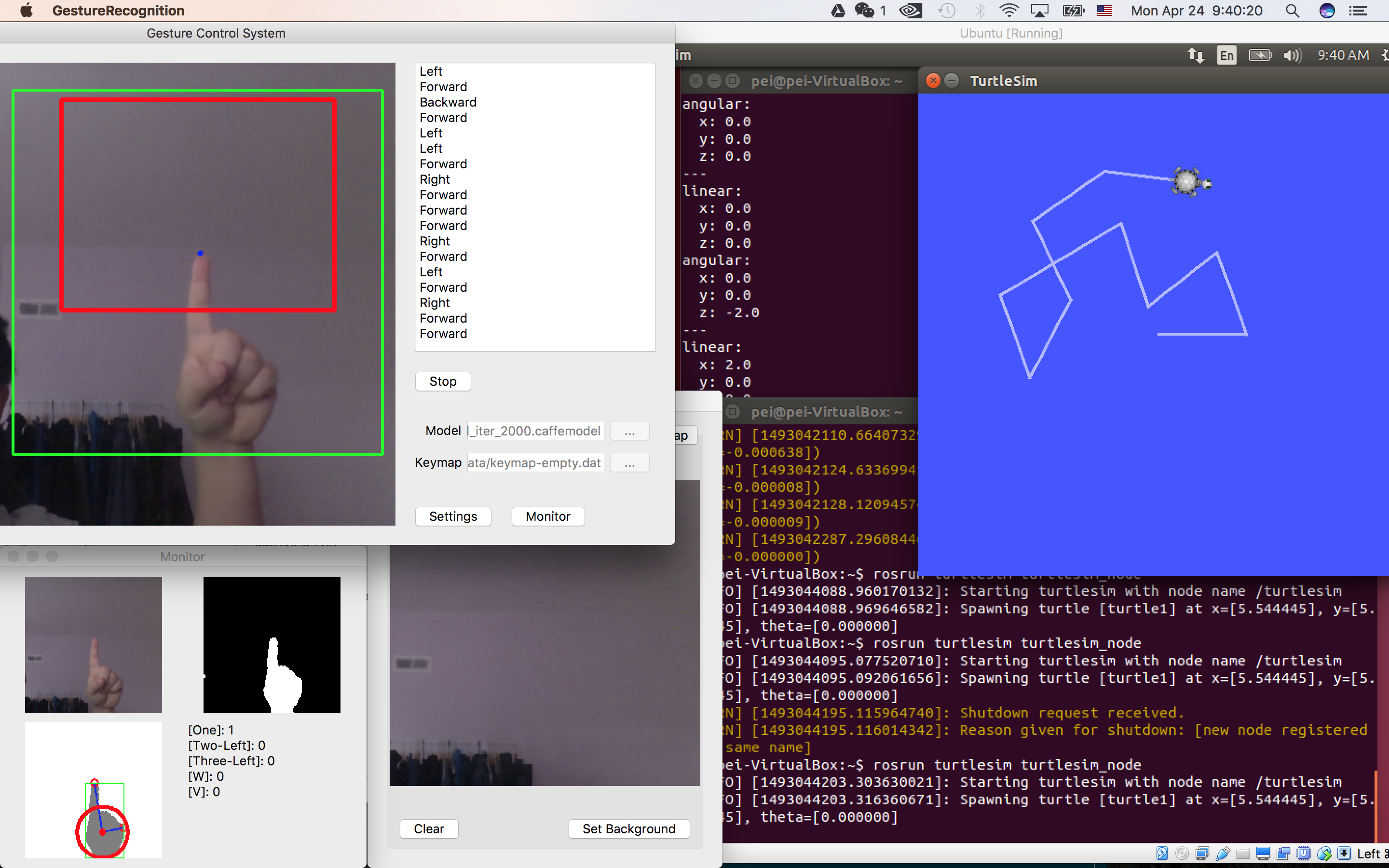}
	\caption{The developed HCI system is extend to the HRI case. The user utilizes gestures to control the movement of a simulated turtle robot through the ROS.}
	\label{fig:robot}
\end{figure}

\subsection{Application in Human-Robot Interaction}
The developed HCI system, in the above experiments, runs to post mouse and keyboard events based on gesture recognition. In most cases, there is no essential difference between posting mouse and keyboard events and sending more complex commands to the computer. We tweak the developed system and make it publish Robot Operating System (ROS) topics when facing certain gestures. Fig.~\ref{fig:robot} displays how we use hand gestures, through the ROS, to control the simulated turtle robot walking. The turtle robot is simulated in a virtual machine, while the developed gesture control system is run on a real operating system. The developed system recognizes gestures and publishes messages. The turtle robot, as a subscriber, receives messages and executing corresponding commands such that the human-robot interaction (HRI) is achieved.

\section{Conclusion}
In this project, we develop a real-time gesture-based HCI system who recognizes gestures only using one monocular camera and extend the system to the HRI case. The developed system relies on a CNN classifier to learn features and to recognize gestures. We employ a series of steps to process the image and to segment the hand region before feeding it to the CNN classifier in order to improve the performance of the CNN classifier. 3,200 gesture images are collected to test the CNN classifier and demonstrate that the CNN classifier combined with our image processing steps can recognize gestures with high accuracy in real time. The usage of the CNN frees us from extracting the gesture features manually and improve the recognition accuracy. Besides, we propose to employ the Kalman filter to smooth the motion of the mouse cursor controlled by the hand, and give a suggestion about how to, on the bare hand, position a point through which to control the movement of the mouse cursor. For the sake of reliability, we, furthermore, propose a simple probabilistic model to effectively prevent the developed system from responding to invalid gestures.

The developed system now only support static gestures. In the future work, we will investigate robust classifiers for dynamic gestures and develop a gesture-based HCI or HRI system with the support of complex motion recognition.


%




\ifCLASSOPTIONcaptionsoff
  \newpage
\fi

\end{document}